\pdfoutput=1

\documentclass[11pt]{article}

\usepackage[final]{acl}

\usepackage{times}
\usepackage{latexsym}
\usepackage{algorithm}
\usepackage{algorithmic}
\usepackage{tabularray}
\usepackage{amsmath}
\DeclareMathOperator*{\argmax}{argmax}

\usepackage[T1]{fontenc}

\usepackage[utf8]{inputenc}

\usepackage{microtype}

\usepackage{inconsolata}

\usepackage{graphicx}

\title{Modeling the One-to-Many Property in Open-Domain Dialogue with LLMs}

\author{Jing Yang Lee\textsuperscript{1}, Kong Aik Lee\textsuperscript{2}, Woon Seng Gan\textsuperscript{3}\\\\
  School of Electrical and Electronic Engineering, Nanyang Technological University\textsuperscript{1,3} \\
  Department of Electrical and Electronic Engineering, The Hong Kong Polytechnic University\textsuperscript{2}\\
  jingyang001@e.ntu.edu.sg\textsuperscript{1}, kong-aik.lee@polyu.edu.hk\textsuperscript{2}, ewsgan@ntu.edu.sg\textsuperscript{3} \\}

\begin{document}
\maketitle
\begin{abstract}
Open-domain Dialogue (OD) exhibits a one-to-many (o2m) property, whereby multiple appropriate responses exist for a single dialogue context. Despite prior research showing that modeling this property boosts response diversity, most modern LLM-based dialogue agents do not explicitly do so. In this work, we model the o2m property of OD in LLMs by decomposing OD generation into two key tasks: Multi-Response Generation (MRG) and Preference-based Selection (PS), which entail generating a set of $n$ semantically and lexically diverse high-quality responses for a given dialogue context, followed by selecting a single response based on human preference, respectively. To facilitate MRG and PS, we introduce o2mDial, a dialogue corpus explicitly designed to capture the o2m property by featuring multiple plausible responses for each context. Leveraging o2mDial, we propose new in-context learning and instruction-tuning strategies, as well as novel evaluation metrics for MRG, alongside a model-based approach for PS. Empirical results demonstrate that applying the proposed two-stage framework to smaller LLMs for OD generation enhances overall response diversity while maintaining contextual coherence, improving response quality by up to 90\%, bringing them closer to the performance of larger models.

\end{abstract}

\section{Introduction}

\begin{figure}[]
    \centering
    \scalebox{0.3}{
    \includegraphics[]{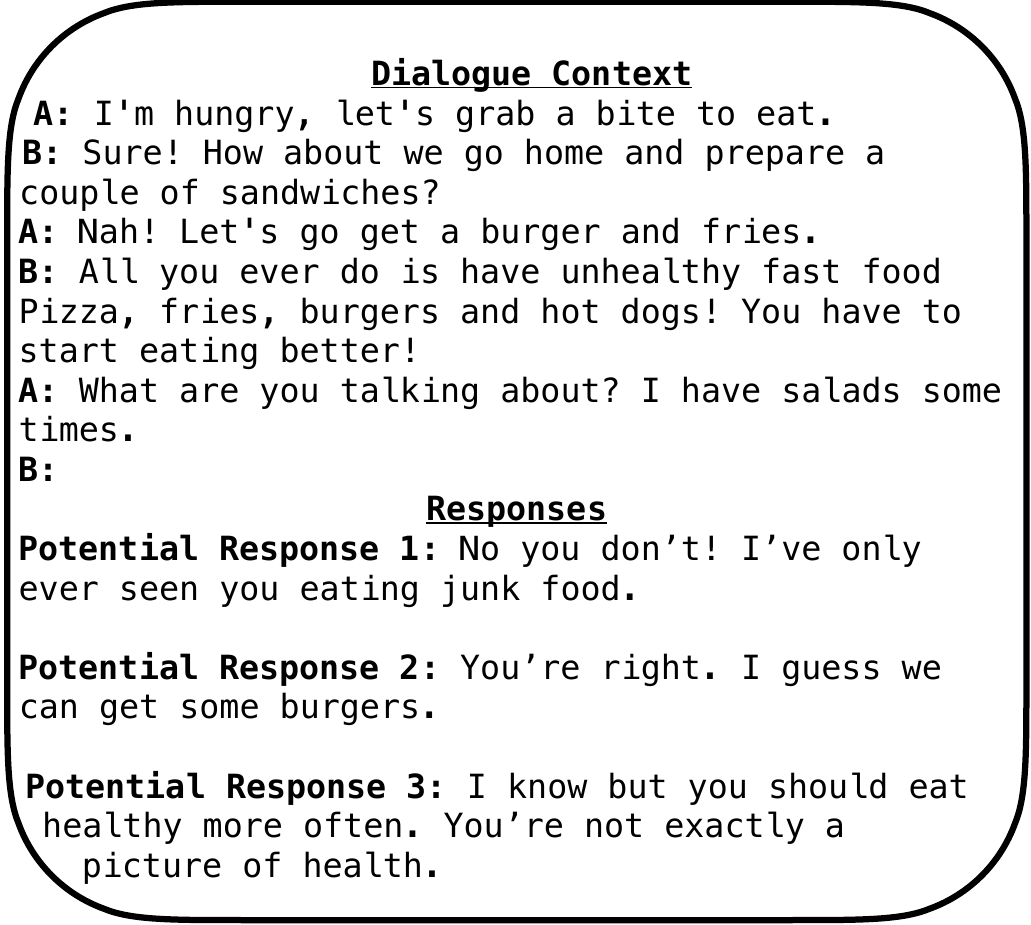}}
    \caption{One-to-many property of open-domain dialogue.}
    \label{fig:o2m}
\end{figure}

Open-domain Dialogue (OD) agents are designed to engage in general conversation across various topics. They aim to generate responses that are fluent, diverse, and contextually coherent with respect to a given dialogue context. Unlike task-oriented agents with specific functions, OD agents simulate human-to-human interaction without predetermined conversational goals. This flexibility leads to the one-to-many (o2m) nature of OD, wherein multiple responses can be derived from a single dialogue context (Figure \ref{fig:o2m}).

Prior research has primarily focused on modeling the o2m property using probabilistic learning frameworks, such as the Conditional Variational Auto-Encoder (CVAE) \cite{shen-etal-2017-conditional, zhao-etal-2017-learning}, to enhance response diverity. These methods typically condition the response on both the dialogue context and a randomly sampled latent variable, capturing the variability in conversational responses and effectively modeling the o2m property. Other approaches include randomized architectures \cite{lee-etal-2022-randomized}, Wasserstein Auto-encoders \cite{gu2018dialogwae}, and Bayesian architectures \cite{lee2023empirical}. These studies illustrate that while explicitly modeling the o2m property of OD significantly enhances response diversity, there is typically a trade-off with contextual coherence \cite{sun-etal-2021-generating, 9747458}.

Recent advancements in Large Language Models (LLMs) have made it increasingly impractical to model the o2m property using probabilistic approaches, primarily due to the immense scale of modern LLMs \cite{zhao2023survey}. These frameworks typically employ a pretrained LLM as the decoder, which is fine-tuned along with additional network components responsible for generating the latent distribution. This process becomes highly resource-intensive given the scale of these LLMs. Moreover, many state-of-the-art LLMs operate as black boxes with undisclosed parameters. Therefore, in the context of LLMs, adopting probabilistic frameworks for generating responses to model the o2m property has become largely impractical.

In this work, instead of adopting a probabilistic approach, we explore modeling the o2m property in LLMs by adopting a two-stage approach by decomposing OD response generation into two subtasks: Multi-Response Generation (MRG) and Preference-based Selection (PS). MRG aims to generate $n$ distinct, contextually coherent responses from a single dialogue context, while PS selects the best response from these $n$ options. For MRG and PS, we introduce o2mDial, a novel dataset designed to capture the o2m property of OD. Each sample in the dataset consists of a dialogue context paired with a set of semantically and lexically distinct yet equally fluent and contextually coherent potential responses. Our two-stage approach focuses on enhancing smaller LLMs ($\leq$ 7 billion parameters), which often face challenges in generating diverse and contextually appropriate responses due to their limited capacity. Empirically, we demonstrate that this approach preserves contextual coherence while significantly increasing response diversity, leading to more engaging interactions with OD dialogue agents, particularly in smaller LLMs. Notably, through automatic and human evaluation, we show that our approach elevates the performance of these smaller models to levels comparable with larger LLMs, which require far greater computational resources. The dataset, metrics, and methodologies introduced in this work provide a valuable resource and baseline for future research into o2m response generation in LLMs.

This paper is organized as follows: MRG and PS are introduced in Section 2 and 3 respectively; Experimental results are provided in Section 4 and Section 5 concludes the paper.

\section{o2mDial} 

To facilitate MRG, we curate o2mDial, a novel conversational dataset that explicitly captures the o2m property of OD. To create o2mDial, we leverage the DailyDialog corpus. First, we sample 500 dialogues (three to six turns) from the training set of the DailyDialog corpus. In this paper, for MRG, we fix $n=5$. In other words, we aim to generate a set of five lexically and semantically distinct, yet contextually coherent responses. Unlike prior datasets that feature multiple reference responses \cite{hedayatnia-etal-2022-systematic, sai-etal-2020-improving, gupta-etal-2019-investigating} that rely on the same LLM to generate every reference response, we use five distinct LLMs to simulate five different agents, with each LLM generating one response. As far as possible, this ensures the semantic and lexical uniqueness of each response. Based on our resource constraints, we selected the following five LLMs: 1)gpt-3.5-turbo \cite{gpt3.5}; 2)llama2-70b-chat \cite{touvron2023llama}; 3)mixtral-8x22b \cite{jiang2024mixtral}; 4) StableVicuna13b \cite{vicuna2023}; 5) Flan-T5-xxl \cite{chung2022scaling}. Additionally, to construct a separate test set for MRG evaluation, we sample another 100 dialogue samples from the test of the DailyDialog corpus. Similar to the training set, each turn consists of a dialogue context (three to six turns), and a set of five distinct and contextually coherent responses.

\begin{figure}[]
    \centering
    \scalebox{0.3}{
    \includegraphics[]{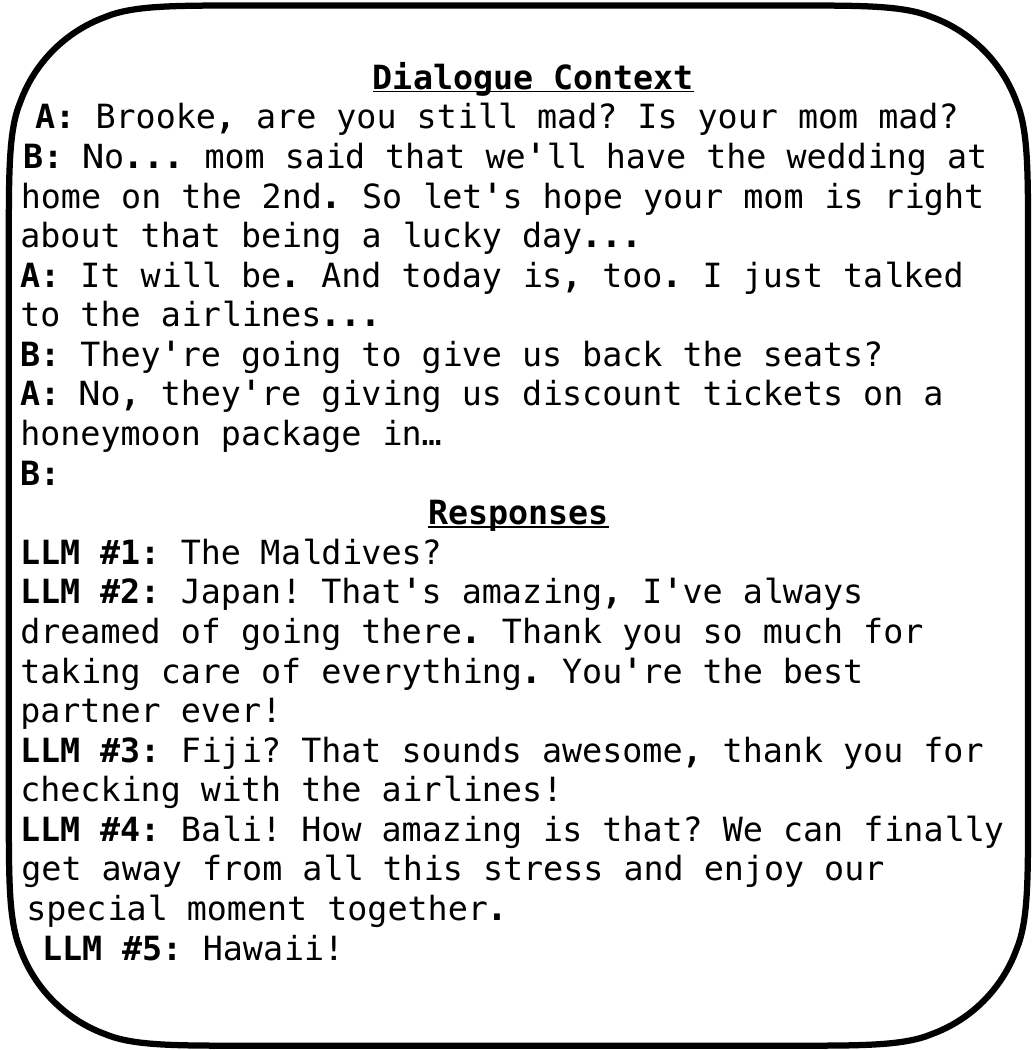}}
    \caption{Sample dialogue context and response set pair from our corpus.}
    \label{fig:corpus_sample}
\end{figure}

Given a dialogue context, each LLM was prompted to generate a one-sentence response. Furthermore, to ensure the quality of our corpus, we manually verify each sample for fluency and contextual coherence. Any responses found to be contextually incoherent or lacking in fluency were manually edited. A sample data point from our corpus in provided in \ref{fig:corpus_sample}. Some statistics regarding the training set of the collected corpus is provided in Table \ref{tbl:corpus_stats}. For PS, we extend o2mDial with additional human preference labels (Section 4). Outside of our two-stage framework, o2mDial could be a useful resource for research for dialogue response evaluation or LLM response preference modeling.

\begin{table}
\centering
\caption{Corpus statistics.}
\label{tbl:corpus_stats}
\scalebox{0.7}{
\begin{tblr}{
  column{2} = {c},
  hline{1,4} = {-}{},
}
\# samples                                              & 500(train)/100(test)\\
Ave \# turns per dialogue context                & 5.3          \\
Ave \# tokens                                 & 14.98 tokens \\
\end{tblr}}
\end{table}

\section{Multi-Response Generation}
\label{sec:MRG}
MRG involves generating a set of $n$ responses given a single dialogue context $D$. In this paper, $R_{n}$ refers to the set of $n$ responses generated by MRG, which consists of utterances alternating between two distinct interlocutors, i.e., $R_{n} = \{r_{1}, r_{2}, \cdots, r_{n}\}$. 

It is vital that responses in $R_{n}$ are contextually coherent as well as lexically and semantically diverse. Semantic diversity requires each response to express a distinct idea, perspective, or piece of information. Lexical diversity involves variations in word choice and structure, allowing responses to differ lexically even if they convey similar ideas. As a result, responses can have high lexical diversity but low semantic diversity. Contextual coherence ensures that responses are logically consistent and relevant to the dialogue context.

It should be noted that even if all $n$ responses convey similar ideas, they can still be lexically unique by using different vocabularies or structures. Thus, a set of responses can exhibit high inter-response lexical diversity while maintaining low inter-response semantic diversity. For example, the statements "A heavy downpour is happening" and "There's a strong rainfall occurring" are lexically unique due to their different words and structures. However, the semantic content—that it is raining heavily—remains the same, indicating high semantic similarity. Our goal is to generate responses with both high inter-response lexical diversity and high inter-response semantic diversity, while ensuring contextual coherence. Prior work in MRG involve utilizing different sampling strategies, which produce responses with low semantic diversity, and pre-specified dialogue acts, which are significantly more complex to implement \cite{sakaeda-kawahara-2022-generate}. 

\subsection{Methods}

In this section, we describe the In-Context Learning (ICL) and Instruction-tuning (IT) approaches we employ for MRG. Unlike prior approaches, we aim to generate $R_{n}$ within a single inference:
\begin{equation}
    R_{n} = LLM(\mathbf{P}(D_{m}))
\end{equation}
where $\mathbf{P}$ refers to a specific prompt template, and $LLM(\cdot)$ denotes any arbitrary LLM. We implement the 3-shot variant of all prompts.

\noindent\textbf{Few-shot (FS) Prompt} This approach involves directly prompting the LLM to generate answers with the task description and demonstrations of query-proactive response pairs. In our experiments, 3 demonstration examples are used. The prompt template is provided in Figure \ref{fig:fs-temp}.

\noindent\textbf{Chain-of-thought (CoT) Prompt} Chain-of-Thought (CoT) prompting \cite{wu-etal-2023-chain} involves prompting the model to generate intermediate steps or explanations in addition to the final answer. In our case, we prompt the LLM to explain how each response differs from the other responses. We hypothesize that by prompting the model to identify the differences between generated responses, the model would be more inclined to generate lexically and semantically diverse responses. The prompt template is provided in Figure \ref{fig:cot-temp}, located in the Appendix.

\noindent\textbf{Prompt Chaining (PC)} Prompt Chaining (PC) \cite{sun-etal-2024-prompt} typically involves dividing a task into smaller subtasks and executing them sequentially using prompts, where the output of one prompt serves as the input for the next. In our approach, we use PC to guide the LLM in generating a set of $n$ unique responses one by one. The process begins with an initial prompt $\mathbf{P}_{0}$ that asks the LLM to generate a response to the dialogue context. Subsequent prompts ($\mathbf{P}_{1} \cdots \mathbf{P}_{n-1}$) instructs the LLM to generate contextually coherent responses that differ semantically and lexically from every response generated by the previous prompts, which are included in the current prompt as input. We hypothesize that by decomposing the task of MRG into $n$ smaller subtasks, the LLM can more effectively ensure both lexical and semantic uniqueness across the responses. However, it is important to note that PC requires multiple inferences from the LLM. Therefore, generating $n$ responses requires $n$ separate inferences, which could impact the feasibility and efficiency of this approach in the real world. The prompt template is provided in Figure \ref{fig:pc-temp}, located in the Appendix.

\noindent\textbf{Demonstration Selection} Furthermore, we perform demonstration selection for the FS, CoT and PC prompts using specific metrics outlined earlier. Specifically, we select responses based on the mean of the inter-response semantic and lexical diversity scores: $sem(R_{n}) + lex(R_{n})$. We identify the top-$k$ responses from our corpus, where $k$ refers to the number of demonstration examples required by the prompt. 


\noindent\textbf{Instruction Tuning (IT)} In addition, we also conduct IT via QLoRA\cite{dettmers2023qlora} using the collected corpus. IT with QLoRA was performed using a batch size of 32, a learning rate of 2e-4, 4 epochs, a rank of 16, an alpha of 32, and a dropout of 0.05. The instruction used for IT is identical to the zero-shot variant (prompt consists of only the instruction without any demonstration examples) of the FS prompt (Figure \ref{fig:fs-temp}).

\subsection{Evaluation}

To measure MRG performance, we design automatic metrics to quantify inter-response semantic and lexical diversity, and overall contextual coherence of $R_{n}$.

\noindent\textbf{Inter-response Semantic \& Lexical Diversity} In the context of open-domain dialogue, response diversity is typically measured via the Distinct metric, which is typically calculated by dividing the number of unique n-grams by the total number of n-grams. However, in our case, we aim to quantify the relative diversity of a set of $n$ responses. In other words, we would like to measure, on average, how different each response is from the other $n-1$ responses. Additionally, based on our definition, it would be ideal if semantic and lexical diversity can be evaluated separately. To this end, we define two separate metrics each accounting for either inter-response semantic or lexical diversity respectively: the inter-response semantic diversity score ($d_{sem}(R_{n})$)  and inter-response lexical diversity score ($d_{lex}(R_{n})$).

For inter-response lexical diversity, we utilize the pairwise edit distance, namely the Jaccard similarity, between every possible response pair in the set:

\begin{equation}
    d_{lex}(R_{n}) = \frac{1}{P_{n}}\sum_{i,j | i \in n, j \in n}\lambda_{Jac}(r_{i}, r_{j})
\end{equation}
where $P_{n}$ refers to the total number of unique pairs in $R_{n}$, $\lambda_{Jac}(\cdot)$ refers to Jaccard similarity. Additionally, on occasion, when a LLM fails to generate the full set of $n$ responses, a value of 1.0 would be assigned as the similarity score for that pair.

For inter-response semantic diversity, we compute the average of the pairwise semantic similarity via the Bert Score among responses in $R_{n}$: 
\begin{equation}
    d_{sem}(R_{n}) = \frac{1}{P_{n}}\sum_{i,j | i \in n, j \in n}\lambda_{BS}(r_{i}, r_{j})
\end{equation}
where $P_{n}$ refers to the total number of unique pairs in $R_{n}$, Likewise, when a LLM fails to generate the full set of $n$ responses, a value of 1.0 would be assigned as the similarity score for that pair. Algorithms for computing $d_{lex}$ and $d_{sem}$ are provided in Algorithm \ref{alg:lexical_similarity} and \ref{alg:semantic_similarity}, respectively.

\noindent\textbf{Contextual Coherence} For our task, the overall contextual coherence of a set of responses can be attained by averaging the individual scores attained by each of the $n$ responses in $R_{n}$ (Algorithm \ref{alg:cc-score}). We employ two contextual coherence metrics: the Utterance Entailment (UE) score \cite{9747458}, and the UniEval-dialog coherence score \cite{zhong-etal-2022-towards}. The UE score involves framing the task of contextual coherence evaluation as a Natural Language Inference (NLI) task. For each utterance in the dialogue context and the corresponding generated response, a NLI model assesses whether the response entails, contradicts, or is neutral with respect to the utterance. For each response $r_{i}$ from $R_{n}$,  $UE(r_{i}) = \frac{1}{m} \sum_{j \in m} NLI(r_{i}, d_{j})$. The UE score of a set of $n$ responses is a continuous number between 0 and 1, where a greater value would indicate greater contextual coherence. The UniEval-dialog is a LLM-based approach which involve re-framing response evaluation as a boolean question and answer task. Essentially a LLM is finetuned and prompted to generate either 'Yes' or 'No' to the question: 'Is this a coherent response given the dialogue history?'. Hence, for the UniEval-dialogue coherence metric, each response is assigned a score of 1 if 'Yes' is generated or 0 if 'No' is generated.


However, evaluating the contextual coherence of OD dialogue responses remains a challenging problem and an active area of research due to the o2m property \cite{li-etal-2016-diversity}. Hence, in our experiments, we conduct a human evaluation to further support our findings.

\section{Preference-based Selection (PS)}
\label{sec:mp-ps}

PS involves selecting the final response $r_{f}$ from $R_{n}$ based on human preference. Unlike traditional open-domain dialogue criteria such as coherence, diversity, engagingness, naturalness, or fluency, human preference covers broader factors like helpfulness, harmlessness, and interestingness \cite{li-etal-2024-dissecting}. We prioritize human preference for three key reasons. Firstly, MRG already ensures coherence and diversity within $R_{n}$. Secondly, existing metrics fall short in capturing the full complexity of human preferences, as they address only specific aspects of response quality \cite{jiang2024surveyhumanpreferencelearning}. Thirdly, modern LLMs are largely capable of generating fluent, natural, and engaging responses \cite{zhao2023survey}. 

However, in addition to human preference, the contextual coherence of the response should still be considered during selection. Hence, for PS, we aim to design an Open-domain Dialogue Response Preference (ODRP) model that assigns a scalar score to each response in $R_{n}$ based on human preference. To achieve this, we leverage an open-source preference model from OpenAssistant \cite{deberta-v3-large-v2} based on deberta-v3-large commonly used for Reinforcement Learning with Human Feedback (RLHF) \cite{bai2022traininghelpfulharmlessassistant}. Such models are typically trained on preference datasets derived from tasks such as summarization \cite{stienon2020learning} and question answering \cite{nakano2021webgpt}, or curated specifically to prevent harmful behavior \cite{bai2022traininghelpfulharmlessassistant}. Hence, to fine-tune the preference model for open-domain dialogue, we construct a new preference dataset from the corpus described earlier.

Preference datasets consist of comparisons between two responses given the same prompt (dialogue context $D$ in our case). Extending o2mDial, we construct a preference dataset for fine-tuning by engaging annotators to label the preferred response $y_{c}$ and the rejected response $y_{r}$ (based on which they would prefer from a conversation partner) for every possible pair of responses from $R_{n}$, resulting in $n \choose 2$ pairs per set. As per \cite{ouyang2022training}, we consider every pair from $R_{n}$ as a single batch. The preference model is then fine-tuned on the following contrastive loss function: 
\begin{equation}
\begin{split}
    J_{\theta} = \frac{1}{{n \choose 2}} E_{(D,y_{c}, y_{r}) \sim R_{n}} [log(\sigma(r(D,y_{c}),r(D,y_{r})))]
\end{split}
\end{equation}
We then proceed to fine-tune the preference model via QLoRA \cite{dettmers2023qlora} for two epochs with AdamW (lr=2e-4). After MRG, the ODRP model assigns a score to each response in $R_{n}$, and $r_{f}$ is the response with the highest score: 

\begin{equation}
    r_f = \argmax_{r \in R_n} ODRP(r)
\end{equation}
Additionally, we introduce a variant of the ODRP model finetuned on a subset of the corpus selected via hard negative sampling \cite{robinson2021contrastive}. Specifically, we apply the base preference model to the dataset and deliberately extracted samples (~50\%) on which the base model performed the worst (assigned a similar score for both $y_{c}$ and $y_{r}$ or assigned a higher score to  $y_{r}$). We finetuned this variant of the ODRP model for four epochs instead.
\section{Experimental Details}
\label{sec:mp-details}
In this section, we outline our experimental design, providing specifics on the corpora utilized, the implementation of our framework, and the baseline approaches employed for comparison.
\subsection{Corpora}
For evaluation, we use two main datasets: DailyDialog \cite{li-etal-2017-dailydialog} and EmpatheticDialogs \cite{rashkin-etal-2019-towards}. DailyDialog features diverse, open-domain multi-turn conversations, while EmpatheticDialogs focuses on responses to emotionally grounded events. In our experiments, the dialogue agent's task is to generate responses based solely on the context of the ongoing conversation. We do not use any additional information such as response labels (e.g., emotion, topic, or style) or speaker labels. 


\subsection{Implementation}
We generate five responses per context ($n=5$) using TinyLlama (v1.1b) \cite{zhang2024tinyllamaopensourcesmalllanguage} and chat variants of Llama2-7b and Llama2-13b \cite{touvron2023llama}. For all experiments, we aim to generate a set of five responses, i.e., $n=5$. The temperature value used in all corpus creation and generation experiments are fixed at 0.7. We do not use other decoding strategies. All experiments were conducted using a single A100 GPU.

\subsection{Baselines}

For MRG, we implement in-context learning via Prompt Chaining (PC) as well as Few-Shot (FS) and Chain-of-Thought (CoT) prompting. We also evaluate Instruction Tuned (IT) variants of the LLM. Additionally, we also generate $R_{n}$ via Multiple Inference (MI). MI entails directly feeding the dialogue context to the LLM and prompting the LLM to generate a single response $n$ times.

For framework evaluation, we utilize PC to generate a response set for each dialogue context in the test set. Subsequently, for PS, we use the fine-tuned ODRP model ($ODRP$) as well as the variant finetuned on hard negative samples ($ODRP$\textsubscript{HN}). Additionally, we introduce the following baseline response selection methods: 1) $rand$: Randomly selecting $r_{f}$ from $R_{n}$; 2) $cls$: Training a classifier (deberta-v2-large) from scratch with the curated preference dataset; 3) $pref$: Using the base OpenAssistant preference model without fine-tuning; 4) base LLM (either $TinyLlama$, $Llama2\text{-}7b$ or $Llama2\text{-}13b$): Generating a response by passing $D$ directly to the LLM i.e., standard LLM inference. Additionally, we leverage the zero-shot variant of the FS prompt (Figure \ref{fig:fs-temp}) to generate a single response from both $Llama2\text{-}70b$ and $gpt\text{-}3.5\text{-}turbo$, allowing us to benchmark these against the responses produced by our framework when implemented with smaller LLMs.

\subsection{Evaluation}

\noindent\textbf{Automatic Evaluation} We evaluate the overall diversity and contextual coherence of the chosen responses by computing the inter-response Distinct-1,2 \cite{li-etal-2016-diversity} and the UE-score \cite{9747458} and UniEval-dialog coherence score \cite{zhong-etal-2022-towards} respectively. To evaluate the set of responses $R_{n}$ generated after MRG, we use several automatic metrics: inter-response semantic diversity ($d_{sem}$) and lexical diversity ($d_{lex}$) scores introduced in Section 3.2, as well as UE-score (UE), and UniEval-dialog coherence score (UniEval) to assess the quality of $R_{n}$.  For inter-response diversity metrics, it should be highlighted that lower scores indicate greater lexical or semantic diversity.

\noindent\textbf{Human Evaluation} In our experiments, we also conduct a human evaluation to evaluate the efficacy of each PS approach. Similar to \cite{smith-etal-2022-human, sakaeda-kawahara-2022-generate}, we engaged a group of five native english speaking participants for a comparative preference-based human evaluation. Each participant was presented a dialogue context along with a response generated by $ODRP$\textsubscript{HN} to compare against each of the other PS approaches ($base$, $rand$, $cls$, $ODRP$), as well as a response generated by $Llama2\text{-}70b$ and $gpt\text{-}3.5\text{-}turbo$, and told to select the agent they would rather converse with. Each participant was presented with 60 samples (30 from DailyDialog and 30 from EmpatheticDialogs) for each comparison. We report the Win, Tie and Loss percentage of each comparison.

In addition, we also conduct a human evaluation to evaluated the quality of the set of responses $R_{n}$ generated during MRG. For this evaluation, we engage a separate group of five native English speakers. Given a set of five responses, each participant was told to count the number of semantically unique responses, lexically unique responses, and contextually coherent responses. Hence, each score is a discrete value from 1 to 5. A count of 5 would imply that all 5 responses were either semantically unique, lexically unique, or contextually coherent. Conversely, a count of 0 would indicate that all 5 responses were semantically similar, lexically similar, or contextually incoherent. Naturally, the participants were not informed which LLM or which generation approach was responsible for each response set. For our generation experiments, each participant was provided with 60 samples (30 from DailyDialog and 30 from EmpatheticDialogs) from each generation approach (the 3 shot variant of each prompt as well as IT, MI, Llama2-70b and gpt-3.5-turbo). Each output consisted of a set of five responses. To illustrate this process, we provide a sample evaluation in Figure \ref{fig:humaneval_sample}, located in the Appendix.


\section{Results \& Discussion}
\label{sec:mp-results}

Here, we assess the performance of the proposed two-stage framework. We also analyze the set responses generated during MRG based on the metrics outlined in Section 3.

\subsection{Framework Evaluation}
The automatic and human evaluation results are presented in Table \ref{tbl:ps-auto} and Table \ref{tbl:ps-human}, respectively. Sample responses are provided in Figure \ref{fig:samples} in the Appendix. 

Based on the results obtained, it is clear that the responses selected by $ODRP$ and $ODRP$\textsubscript{HN} consistently outperform all other approaches, including $rand$, $cls$, and $pref$, in terms of both diversity and contextual coherence. Both $ODRP$ and $ODRP\textsubscript{HN}$ generally achieve statistically significantly higher Distinct and UE/UniEval scores than the baseline methods. Moreover, in human evaluation, they show a greater proportion of wins and a lower proportion of losses compared to other baselines. Qualitatively, we observe that responses selected by $ODRP$ and $ODRP$\textsubscript{HN} do more than just acknowledge the previous utterance; they often provide additional enriching information that enhances the overall dialogue. Furthermore, a significant portion of these selected responses include queries directed at the other interlocutor, actively encouraging further interaction. 

It is also important to note that fine-tuning the ODRP model with hard negative samples leads to a noticeable improvement in the diversity and coherence of the selected responses across all LLMs. $ODRP$\textsubscript{HN} outperforms $ODRP$ on all automatic metrics and achieves a higher Win rate and lower Loss rate in human evaluation. The effectiveness of the ODRP model is particularly evident in the case of TinyLlama, where there is substantial variability in the quality of responses generated  during MRG. Generally, we observe that the ODRP model excels at identifying and prioritizing higher-quality responses, resulting in more engaging and meaningful exchanges, even when the initial set of responses exhibits significant variability. This leads to improvements of up to 90\% in response diversity and contextual coherence.

\noindent\textbf{Comparison with Larger LLMs} In addition, we evaluated larger LLMs, such as Llama2-70b and gpt-3.5-turbo, using the zero-shot variant of the FS prompt (Figure \ref{fig:fs-temp}). Our findings reveal that after applying our two-stage framework and selecting responses via $ODRP\textsubscript{HN}$, the quality of responses generated by smaller LLMs like $TinyLlama$ and $Llama2\text{-}7b$ surpasses that of $Llama2\text{-}70b$ in terms of response diversity and approaches the level of $gpt\text{-}3.5\text{-}turbo$. Regarding contextual coherence, $Llama2\text{-}13b$ see improvements that bring it in line with $Llama2\text{-}70b$ and $gpt\text{-}3.5\text{-}turbo$, while $TinyLlama$ and $Llama2\text{-}7b$, although still trailing, narrow the gap significantly. Qualitatively, we note that responses selected by $ODRP$\textsubscript{HN} are comparable to responses generated by $Llama2\text{-}70b$ and $gpt\text{-}3.5\text{-}turbo$ in terms of naturalness and engagingness. These results underscore the effectiveness of our approach, enabling smaller LLMs to rival or exceed the capabilities of larger models, all while maintaining lower computational demands.

\subsection{MRG Evaluation} 

In addition, we evaluate the MRG performance of 3-shot FS, CoT, PC, and IT on the o2mDial test set. Automatic and human evaluation results are presented in Table \ref{tbl:eval_test}.

We observe that larger LLMs like Llama2-7b and 13b generally outperform TinyLlama, likely due to their superior instruction-following abilities, which enhance in-context learning and IT effectiveness. The PC and IT methods yield results comparable to reference responses in the test set for Llama2-7b and 13b, while TinyLlama lags slightly, reflecting its weaker capabilities. Despite TinyLlama’s limitations, PC’s simpler task breakdown marginally improved performance, outperfroming all other baseline MRG methods. Llama2-7b and 13b also benefited from PC and CoT prompts, boosting response diversity while preserving contextual coherence, as shown by comparable UE/UniEval scores.

Closer examination of the responses reveal that quality rises with model size—TinyLlama produces the weakest outputs, while Llama2-13b excels. All three models faced issues: insufficient responses (below $n$), redundancy (similar or identical replies), and over-extended conversations (too many utterances). Insufficient and redundant responses reduced semantic and lexical diversity, while over-extensions impacted coherence metrics like UE and UniEval scores. TinyLlama had more insufficient responses, Llama2-7b and 13b saw occasional over-extensions, and redundancy appeared across all models, most prominently in TinyLlama. Generally, there remains a performance gap between the reference responses and proposed approaches. Future work will aim to reduce this gap.


\noindent\textbf{Comparison with MI} Response sets generated via MI tend to be semantically similar despite relatively high lexical diversity, as shown by low inter-response semantic scores and comparably higher lexical diversity scores in both automatic and human evaluations. This is likely due to the deterministic nature of logits during inference. Although sampling strategies (temperature scaling \cite{10.5555/3305381.3305518} or nucleus sampling \cite{holtzman2020curiouscaseneuraltext}) introduce stochasticity in decoding, generated logits remain deterministic, limiting semantic variation unless randomness is significantly increased, which could reduce contextual coherence.

\section{Related Work}


Prior work adopting a two-stage approach for open-domain dialogue typically involves generating multiple responses either through conditional generation based on pre-specified dialogue acts \cite{sakaeda-kawahara-2022-generate} or by pooling outputs from variational and retrieval-based systems \cite{10.1016/j.csl.2020.101071, Technology2021}. However, these studies often focus on evaluating only the final selected response, without considering the diversity or contextual coherence of the entire set of generated responses. In contrast, our approach evaluates and optimizes the quality of the full set of responses, thereby enhancing the overall quality of the final selected response. Additionally, many of these methods have been applied to smaller language models, whereas to the best of our knowledge, our work is the first to introduce a two-stage generation framework LLMs. Other two-stage approaches broadly entail first generating a candidate response and instantiating it as the final response \cite{10.1145/3551869}, or generating a response in the first stage and further conditioning and refining the response in the second stage \cite{qian2024thinktwicehumanliketwostage, 9525043}.

Regarding response selection, prior work has primarily concentrated on narrow criteria such as engagement \cite{sakaeda-kawahara-2022-generate}, topical relevance \cite{10.1016/j.csl.2020.101071, YUAN2024100087}. Standard retrieval-based systems, in contrast, prioritize contextual coherence \cite{ijcai2021p627, su-etal-2024-dial}. In our framework, we prioritize human preferences, considering a broader range of factors such as harmlessness and helpfulness, which are critical aspects for ensuring the real-world utility of response generation systems.

\begin{table}
\centering
\caption{Automatic evaluation results. The best score in each column is \textbf{bolded}. * indicates a statistically significant difference in score (t-test, $p$-value \textless 0.01) from the \textbf{bolded} score. Scores for DailyDialog and EmpatheticDialogues are provided before and after the backslash '\symbol{92}', respectively.}
\label{tbl:ps-auto}
\scalebox{0.7}{
\begin{tblr}{
  cell{1}{2} = {c},
  cell{1}{3} = {c},
  cell{1}{4} = {c},
  cell{1}{5} = {c},
  cell{2}{2} = {c},
  cell{2}{3} = {c},
  cell{2}{4} = {c},
  cell{2}{5} = {c},
  cell{3}{2} = {c},
  cell{3}{3} = {c},
  cell{3}{4} = {c},
  cell{3}{5} = {c},
  cell{4}{2} = {c},
  cell{4}{3} = {c},
  cell{4}{4} = {c},
  cell{4}{5} = {c},
  cell{5}{2} = {c},
  cell{5}{3} = {c},
  cell{5}{4} = {c},
  cell{5}{5} = {c},
  cell{6}{2} = {c},
  cell{6}{3} = {c},
  cell{6}{4} = {c},
  cell{6}{5} = {c},
  cell{7}{2} = {c},
  cell{7}{3} = {c},
  cell{7}{4} = {c},
  cell{7}{5} = {c},
  cell{8}{2} = {c},
  cell{8}{3} = {c},
  cell{8}{4} = {c},
  cell{8}{5} = {c},
  cell{9}{2} = {c},
  cell{9}{3} = {c},
  cell{9}{4} = {c},
  cell{9}{5} = {c},
  cell{10}{2} = {c},
  cell{10}{3} = {c},
  cell{10}{4} = {c},
  cell{10}{5} = {c},
  cell{11}{2} = {c},
  cell{11}{3} = {c},
  cell{11}{4} = {c},
  cell{11}{5} = {c},
  cell{12}{2} = {c},
  cell{12}{3} = {c},
  cell{12}{4} = {c},
  cell{12}{5} = {c},
  cell{13}{2} = {c},
  cell{13}{3} = {c},
  cell{13}{4} = {c},
  cell{13}{5} = {c},
  cell{14}{2} = {c},
  cell{14}{3} = {c},
  cell{14}{4} = {c},
  cell{14}{5} = {c},
  cell{15}{2} = {c},
  cell{15}{3} = {c},
  cell{15}{4} = {c},
  cell{15}{5} = {c},
  cell{16}{2} = {c},
  cell{16}{3} = {c},
  cell{16}{4} = {c},
  cell{16}{5} = {c},
  cell{17}{2} = {c},
  cell{17}{3} = {c},
  cell{17}{4} = {c},
  cell{17}{5} = {c},
  cell{18}{2} = {c},
  cell{18}{3} = {c},
  cell{18}{4} = {c},
  cell{18}{5} = {c},
  cell{19}{2} = {c},
  cell{19}{3} = {c},
  cell{19}{4} = {c},
  cell{19}{5} = {c},
  hline{1-2,8,14,20,22} = {-}{},
  rowsep=1.0pt
}
              & Dist-1                      & Dist-2                      & UE                          & UniEval                     \\
$TinyLlama$     & 0.16*/0.18*                 & 0.51*/0.61*                 & 0.21*/0.13*                 & 0.74*/0.64*                 \\
- $rand$        & 0.24*/0.20*                 & 0.75*/0.70*                 & 0.24/0.13*                  & 0.76*/0.65*                 \\
- $cls$         & 0.22*/0.25*                 & 0.76*/0.74*                 & 0.23/0.18*                  & 0.78*/0.66*                 \\
- $pref$        & 0.25*/0.24*                 & 0.73*/0.75*                 & 0.24/0.18*                  & 0.77*/0.70*                 \\
- $ODRP$        & 0.28*/0.29                  & 0.77/0.798                  & 0.27/0.22*                  & 0.81/0.72*                  \\
- $ODRP$\textsubscript{HN}      & \textbf{0.31/0.31}          & \textbf{0.79/0.82}          & \textbf{0.30/0.26}          & \textbf{0.83/0.76}          \\
$Llama2\text{-}7b$     & 0.20*/0.22*                 & 0.61*/0.69*                 & 0.24*/0.21                  & 0.83/0.72                   \\
- $rand$~       & 0.23*/0.30*                 & 0.77*/0.78*                 & 0.22*/0.19*                 & 0.81*/0.69*                 \\
- $cls$         & 0.30*/0.27*                 & 0.79*/0.75*                 & 0.23*/0.18*                 & 0.80*/0.72                  \\
- $pref$        & 0.28/0.29*                  & 0.77*/0.78*                 & 0.24*/0.22                  & 0.83/0.71*                  \\
- $ODRP$        & 0.33/0.35                   & \textbf{0.83}/0.84          & 0.26/0.22                   & 0.83/\textbf{0.73}          \\
- $ODRP$\textsubscript{HN}     & \textbf{0.35/0.36}          & \textbf{0.83}/\textbf{0.85} & \textbf{0.29}/\textbf{0.24} & \textbf{0.85}/\textbf{0.73} \\
$Llama2\text{-}13b$    & 0.21*/0.23*                 & 0.65*/0.72*                 & 0.26*/0.24*                 & 0.85/0.77*                  \\
- $rand$~       & 0.24*/0.28*                 & 0.77*/0.76*                 & 0.25*/0.24*                 & 0.80*/0.72*                 \\
- $cls$         & 0.30*/0.31*                 & 0.80*/0.76*                 & 0.29*/0.25                  & 0.83*/0.77*                 \\
- $pref$        & 0.31/0.30*                  & 0.79*/0.78                  & 0.26*/0.29                  & 0.82*/0.79                  \\
- $ODRP$        & \textbf{0.33}/0.34          & \textbf{0.85}/0.79          & 0.32/0.30                   & 0.85/0.81                   \\
- $ODRP$\textsubscript{HN}     & \textbf{0.33}/\textbf{0.35} & 0.84/\textbf{0.82}          & \textbf{0.33}/\textbf{0.32} & \textbf{0.87}/\textbf{0.82} \\
$Llama2\text{-}70b$    & 0.31/0.32                   & 0.72/0.80                   & 0.28/0.26                   & 0.86/0.79                   \\
$gpt\text{-}3.5\text{-}turbo$ & 0.36/0.33                   & 0.75/0.82                   & 0.31/0.30                   & 0.88/0.81                   
\end{tblr}}
\end{table}

\begin{table}
\centering
\caption{Human evaluation results. The Win, Tie, and Loss percentages are presented for each comparison.}
\label{tbl:ps-human}
\scalebox{0.7}{
\begin{tblr}{
  column{3} = {c},
  column{4} = {c},
  column{5} = {c},
  cell{2}{1} = {r=7}{},
  cell{9}{1} = {r=7}{},
  cell{16}{1} = {r=7}{},
  hline{1-2,9,16,23} = {-}{},
  rowsep=1.0pt
}
           &                          & Win & Tie & Loss\\
TinyLlama  & $ODRP$\textsubscript{HN} vs. $TinyLlama$          & 85      & 9       & 6        \\
           & $ODRP$\textsubscript{HN} vs. $rand$          & 76      & 16      & 18       \\
           & $ODRP$\textsubscript{HN} vs. $cls$           & 60      & 29      & 11       \\
           & $ODRP$\textsubscript{HN} vs. $pref$          & 57      & 20      & 23       \\
           & $ODRP$\textsubscript{HN} vs. $ODRP$          & 49      & 33      & 18       \\
           & $ODRP$\textsubscript{HN} vs. $Llama2\text{-}70b$    & 30      & 35      & 35       \\
           & $ODRP$\textsubscript{HN} vs. $gpt\text{-}3.5\text{-}turbo$ & 26      & 44      & 30       \\
Llama2-7b  & $ODRP$\textsubscript{HN} vs. $Llama2\text{-}7b$          & 74      & 18      & 8        \\
           & $ODRP$\textsubscript{HN} vs. $rand$          & 58      & 25      & 17       \\
           & $ODRP$\textsubscript{HN} vs. $cls$           & 50      & 29      & 21       \\
           & $ODRP$\textsubscript{HN} vs. $pref$          & 47      & 27      & 23       \\
           & $ODRP$\textsubscript{HN} vs. $ODRP$          & 46      & 30      & 24       \\
           & $ODRP$\textsubscript{HN} vs. $Llama2\text{-}70b$    & 32      & 41      & 27       \\
           & $ODRP$\textsubscript{HN} vs. $gpt\text{-}3.5\text{-}turbo$ & 28      & 48      & 24       \\
Llama2-13b & $ODRP$\textsubscript{HN} vs. $Llama2\text{-}13b$          & 50      & 33      & 17       \\
           & $ODRP$\textsubscript{HN} vs. $rand$          & 51      & 24      & 25       \\
           & $ODRP$\textsubscript{HN} vs. $cls$           & 44      & 34      & 22       \\
           & $ODRP$\textsubscript{HN} vs. $pref$          & 42      & 30      & 28       \\
           & $ODRP$\textsubscript{HN} vs. $ODRP$          & 41      & 32      & 27       \\
           & $ODRP$\textsubscript{HN} vs. $Llama2\text{-}70b$    & 38      & 39      & 23       \\
           & $ODRP$\textsubscript{HN} vs. $gpt\text{-}3.5\text{-}turbo$ & 37      & 40      & 23       
\end{tblr}}
\end{table}

\begin{table}
\centering
\caption{MRG automatic and human evaluation results on the o2mDial test set.}
\label{tbl:eval_test}
\scalebox{0.7}{
\begin{tblr}{
  row{1} = {c},
  row{3} = {c},
  row{4} = {c},
  row{5} = {c},
  row{6} = {c},
  row{8} = {c},
  row{9} = {c},
  row{10} = {c},
  row{11} = {c},
  row{13} = {c},
  row{14} = {c},
  row{15} = {c},
  row{16} = {c},
  row{17} = {c},
  row{20} = {c},
  row{21} = {c},
  row{22} = {c},
  row{23} = {c},
  row{25} = {c},
  row{26} = {c},
  row{27} = {c},
  row{28} = {c},
  row{30} = {c},
  row{31} = {c},
  row{32} = {c},
  row{33} = {c},
  row{34} = {c},
  cell{2}{1} = {r=5}{},
  cell{2}{2} = {c},
  cell{2}{3} = {c},
  cell{2}{4} = {c},
  cell{2}{5} = {c},
  cell{2}{6} = {c},
  cell{7}{1} = {r=5}{},
  cell{7}{2} = {c},
  cell{7}{3} = {c},
  cell{7}{4} = {c},
  cell{7}{5} = {c},
  cell{7}{6} = {c},
  cell{12}{1} = {r=5}{},
  cell{12}{2} = {c},
  cell{12}{3} = {c},
  cell{12}{4} = {c},
  cell{12}{5} = {c},
  cell{12}{6} = {c},
  cell{17}{1} = {c=2}{},
  cell{18}{1} = {c},
  cell{18}{3} = {c},
  cell{18}{4} = {c},
  cell{18}{5} = {c},
  cell{18}{6} = {c},
  cell{19}{1} = {r=5}{},
  cell{19}{2} = {c},
  cell{19}{3} = {c},
  cell{19}{4} = {c},
  cell{19}{5} = {c},
  cell{19}{6} = {c},
  cell{24}{1} = {r=5}{},
  cell{24}{2} = {c},
  cell{24}{3} = {c},
  cell{24}{4} = {c},
  cell{24}{5} = {c},
  cell{24}{6} = {c},
  cell{29}{1} = {r=5}{},
  cell{29}{2} = {c},
  cell{29}{3} = {c},
  cell{29}{4} = {c},
  cell{29}{5} = {c},
  cell{29}{6} = {c},
  cell{34}{1} = {c=2}{},
  hline{1-2,7,12,17-19,24,29,34-35} = {-}{},
    rowsep=1.0pt
}
Model      &     & $d\textsubscript{sem}$ & $d\textsubscript{lex}$ & UE            & UniEval       \\
TinyLlama  & MI  & 0.86                   & 0.78                   & 0.20          & 0.73          \\
           & FS  & 0.66*                  & 0.75*                  & 0.19*         & 0.72*         \\
           & CoT & 0.67*                  & 0.74*                  & 0.21*         & 0.74*         \\
           & PC  & 0.64                   & 0.70*                  & 0.25*         & 0.77*         \\
           & IT  & 0.65*                  & 0.72*                  & 0.23*         & 0.75*         \\
Llama2-7b  & MI  & 0.81                   & 0.76                   & 0.24          & 0.82          \\
           & FS  & 0.65*                  & 0.74*                  & 0.25*         & 0.80*         \\
           & CoT & 0.62                   & 0.67*                  & 0.28*         & 0.86          \\
           & PC  & 0.60                   & 0.65*                  & 0.28*         & 0.87          \\
           & IT  & 0.65*                  & 0.68*                  & 0.26*         & 0.84*         \\
Llama2-13b & MI  & 0.74                   & 0.70                   & 0.28          & 0.84          \\
           & FS  & 0.61                   & 0.68*                  & 0.29*         & 0.85*         \\
           & CoT & \textbf{0.60}          & 0.65*                  & 0.28*         & 0.88          \\
           & PC  & \textbf{0.60}          & 0.66*                  & 0.30          & 0.88          \\
           & IT  & 0.61                   & 0.67*                  & 0.29*         & 0.87          \\
Reference  &     & \textbf{0.60}          & \textbf{0.62}          & \textbf{0.32} & \textbf{0.89} \\
           &     & Sem. Div.              & Lex. Div.              & Con. Coh.     & $\kappa$      \\
TinyLlama  & MI  & 1.89                   & 1.95                   & 3.95          & 0.54          \\
           & FS  & 3.42                   & 3.82                   & 3.95          & 0.55          \\
           & CoT & 3.58                   & 3.88                   & 3.91          & 0.54          \\
           & PC  & 3.70                   & 3.96                   & 3.98          & 0.51          \\
           & IT  & 3.75                   & 4.01                   & 3.99          & 0.49          \\
Llama2-7b  & MI  & 2.33                   & 2.45                   & 4.73          & 0.58          \\
           & FS  & 4.30                   & 4.60                   & 4.79          & 0.57          \\
           & CoT & 4.44                   & 4.72                   & 4.85          & 0.59          \\
           & PC  & 4.58                   & 4.73                   & 4.85          & 0.66          \\
           & IT  & 4.53                   & 4.70                   & 4.70          & 0.60          \\
Llama2-13b & MI  & 2.67                   & 2.92                   & 4.88          & 0.47          \\
           & FS  & 4.44                   & 4.66                   & 4.82          & 0.50          \\
           & CoT & 4.65                   & 4.74                   & 4.88          & 0.58          \\
           & PC  & 4.66                   & 4.71                   & 4.89          & \textbf{0.59} \\
           & IT  & 4.55                   & 4.69                   & 4.80          & 0.54          \\
Reference  &     & \textbf{4.69}          & \textbf{4.77}          & \textbf{4.89} & 0.58          
\end{tblr}}
\end{table}

\section{Conclusion}
\label{sec:mp-con}

This paper decomposes OD response generation into Multi-Response Generation (MRG) and Preference-based Selection (PS). For MRG, we curate o2mDial and propose methods such as FS, CoT, PC, and IT. We also introduce metrics to evaluate semantic and lexical diversity. For PS, we develop the ODRP model to select responses aligned with human preferences. Empirical results show MRG and PS significantly enhance response diversity by up to 90\% in smaller LLMs, achieving performance on par with larger LLMs. Future research could expand the number of unique responses per set (beyond $n = 5$) to assess impacts on diversity and quality. Systematically increasing $n$ could help identify the optimal point of diminishing returns. For PS, another potential avenue for additional research could involve integrating dialogue context into the evaluation process to act as a safeguard against contextually incoherent responses.

\section{Limitations}
Due to resource limitations, the LLMs employed for dataset curation in our experiments are intentionally smaller in size. Future work could entail extending o2mDial with larger, more recent LLMs. Furthermore, due to time and resource constraints, exhaustive prompt engineering was not performed for each model. Instead, we focused on basic prompt engineering techniques aimed at ensuring consistent and coherent output formatting. While this approach was sufficient for the scope of the experiments, we acknowledge that more sophisticated and fine-tuned prompt engineering could potentially improve the models' performance in more complex or specialized tasks.
\bibliography{custom}
\clearpage
\appendix

\onecolumn
\section{Appendix}
\label{sec:appendix}

\begin{algorithm*}
\caption{Inter-response lexical similarity score $d_{lex}$.}
\label{alg:lexical_similarity}
\begin{algorithmic}
\REQUIRE Set of $n$ responses $R_{n}$, Jaccard Similarity function $J(\cdot)$
\ENSURE Lexical similarity score $s$

\STATE $s_{t} \gets 0 $ \COMMENT{Initialize temporary score}
\STATE $P \gets 0 $ \COMMENT{Initialize pair count}

\FOR{$i \gets 0$ to $n-1$}
    \FOR{$j \gets i + 1$ to $n-1$}
        \IF{$r_{i} = \text{None}$ \textbf{or} $r_{j} = \text{None}$}
            \STATE $s_{t} \gets s_{t} + 1.0$
        \ELSE
            \STATE $s_{t} \gets s_{t} + \lambda_{Jac}(r_{i}, r_{j})$
        \ENDIF
        \STATE $P \gets P + 1 $ \COMMENT{Increment pair count}
    \ENDFOR
\ENDFOR

\STATE $s \gets \frac{1}{P} s_{t}$ \COMMENT{Compute mean over all pairs}
\RETURN $s$

\end{algorithmic}
\end{algorithm*}

\begin{algorithm*}
\caption{Inter-response semantic similarity score $d_{sem}$.}
\label{alg:semantic_similarity}
\begin{algorithmic}
\REQUIRE Set of $n$ responses $R_{n}$, BertScore function $BS(\cdot)$
\ENSURE Semantic similarity score $s$

\STATE $s_{t} \gets 0 $ \COMMENT{Initialize temporary score}
\STATE $P \gets 0 $ \COMMENT{Initialize pair count}

\FOR{$i \gets 0$ to $n-1$}
    \FOR{$j \gets i + 1$ to $n-1$}
        \IF{$r_{i} = \text{None}$ \textbf{or} $r_{j} = \text{None}$}
            \STATE $s_{t} \gets s_{t} + 1.0$
        \ELSE
            \STATE $s_{t} \gets s_{t} + \lambda_{BS}(r_{i}, r_{j})$
        \ENDIF
        \STATE $P \gets P + 1$ \COMMENT{Increment pair count}
    \ENDFOR
\ENDFOR

\STATE $s \gets \frac{1}{P} s_{t}$ \COMMENT{Compute mean over all pairs}
\RETURN $s$

\end{algorithmic}
\end{algorithm*}

\begin{algorithm*}
\caption{Contextual Coherence score}
\label{alg:cc-score}
\begin{algorithmic}
\REQUIRE Set of $n$ responses $R_{n}$, set of $m$ dialogue context $D_{m}$, Contextual Coherence measure $CC(\cdot)$ (e.g., UE score or UniEval-dialogue coherence score)
\ENSURE Contextual coherence score $s$

\STATE $s_{t} \gets 0 $ \COMMENT{Initialize temporary score}

\FOR{$i \gets 0$ to $n-1$}
    \IF{$r_{i} = \text{None}$}
        \STATE $s_{t} \gets s_{t} + 0.0$
    \ELSE
        \STATE $s_{t} \gets s_{t} + CC(r_{i}, D_{m})$
    \ENDIF
\ENDFOR

\STATE $s \gets \frac{1}{n} s_{t}$ \COMMENT{Compute mean over $n$ responses}
\RETURN $s$

\end{algorithmic}
\end{algorithm*}

\begin{figure*}[]
    \centering
    \scalebox{0.4}{
    \includegraphics[]{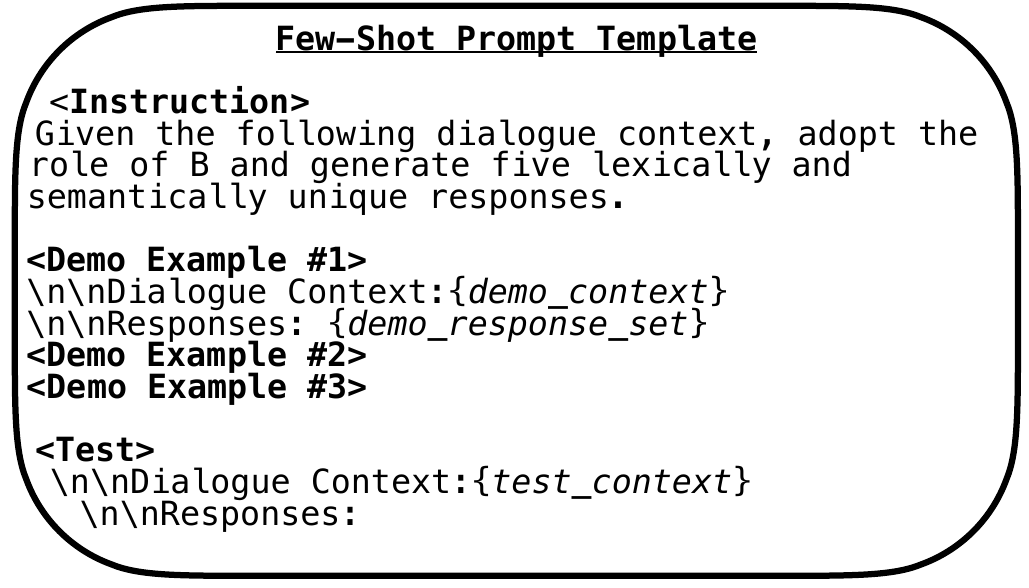}}
    \caption{Prompt template for the Few-Shot prompt.}
    \label{fig:fs-temp}
\end{figure*}

\begin{figure*}[]
    \centering
    \scalebox{0.4}{
    \includegraphics[]{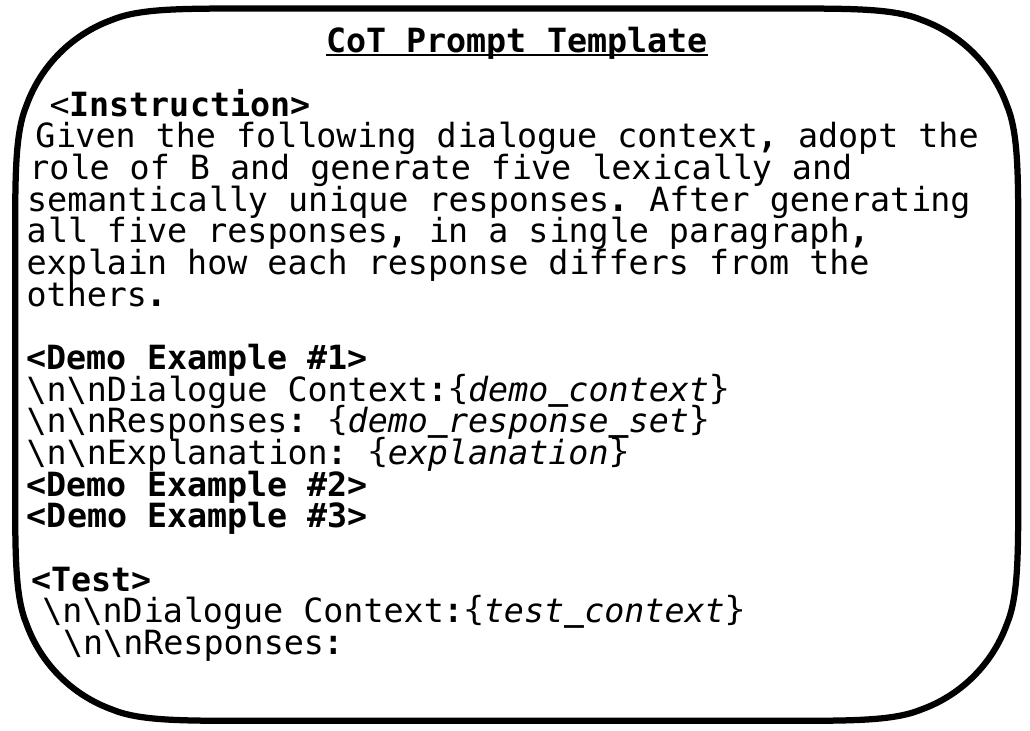}}
    \caption{Prompt template for the Chain-of-Thought prompt.}
    \label{fig:cot-temp}
\end{figure*}

\begin{figure*}[]
    \centering
    \scalebox{0.4}{
    \includegraphics[]{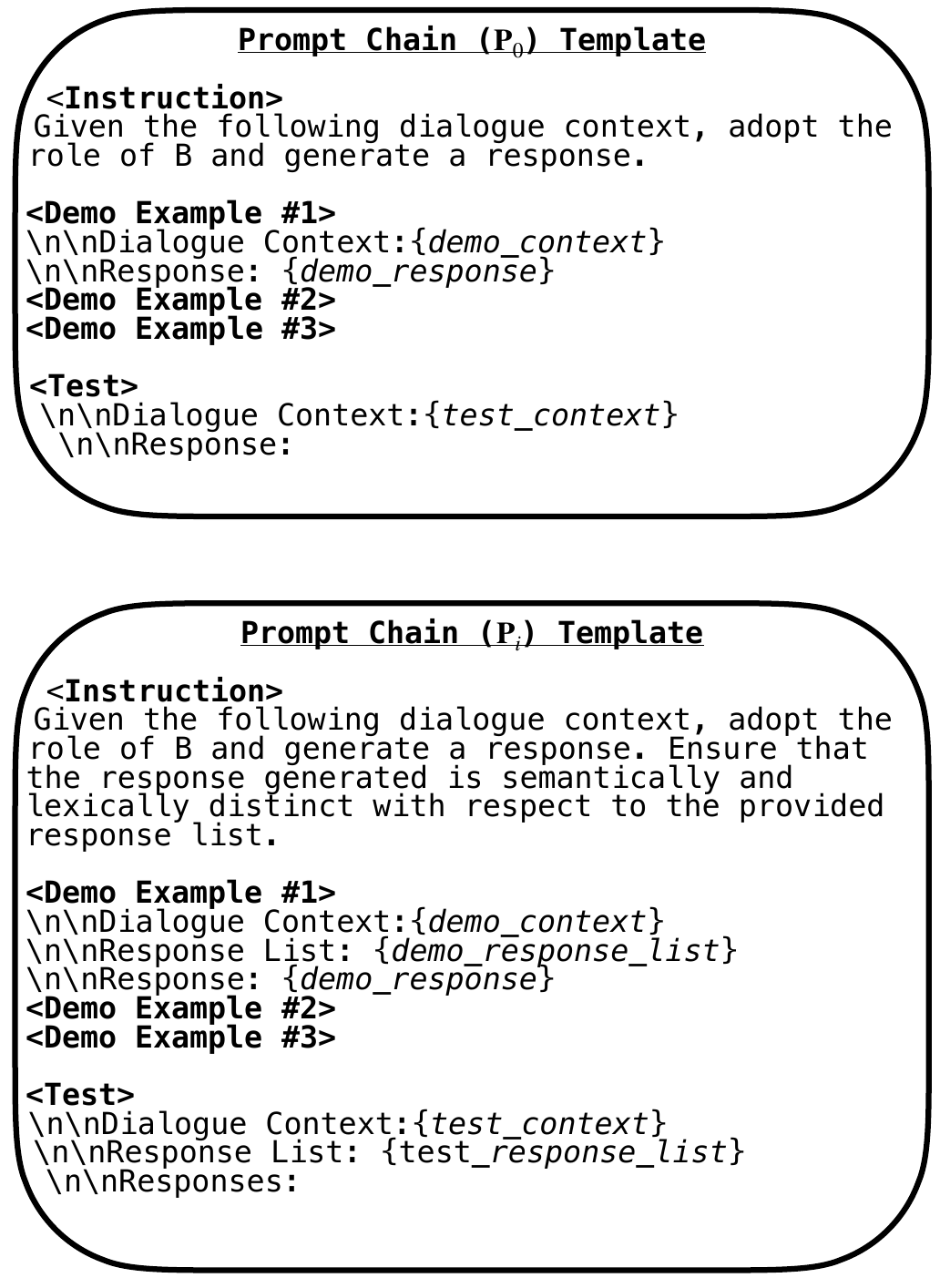}}
    \caption{Prompt template for the Prompt Chain (PC).}
    \label{fig:pc-temp}
\end{figure*}


\begin{figure*}[]
    \centering
    \scalebox{0.4}{
    \includegraphics[]{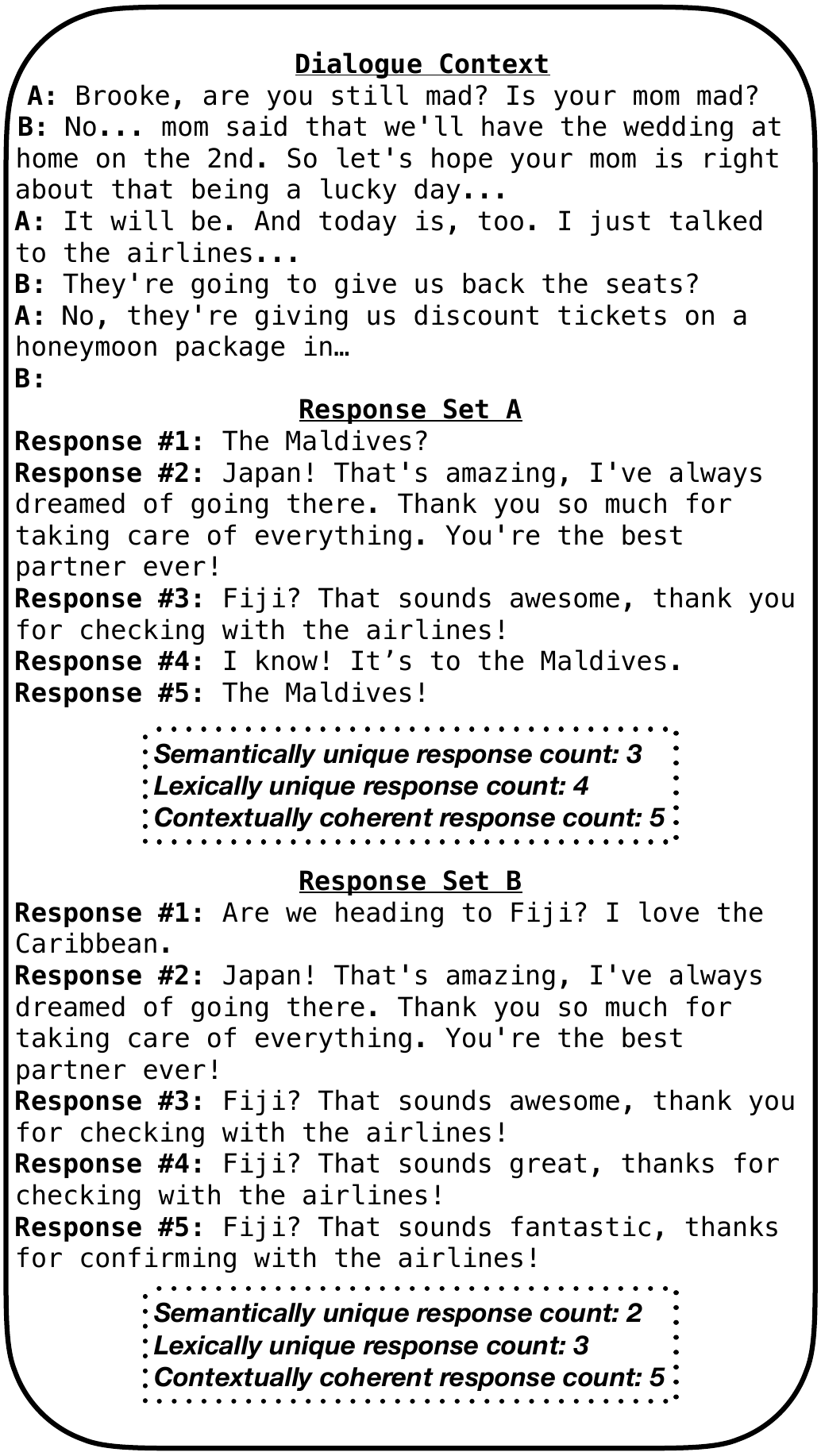}}
    \caption{Samples of human evaluation used for MRG.}
    \label{fig:humaneval_sample}
\end{figure*}

\begin{figure*}[]
    \centering
    \scalebox{0.4}{
    \includegraphics[]{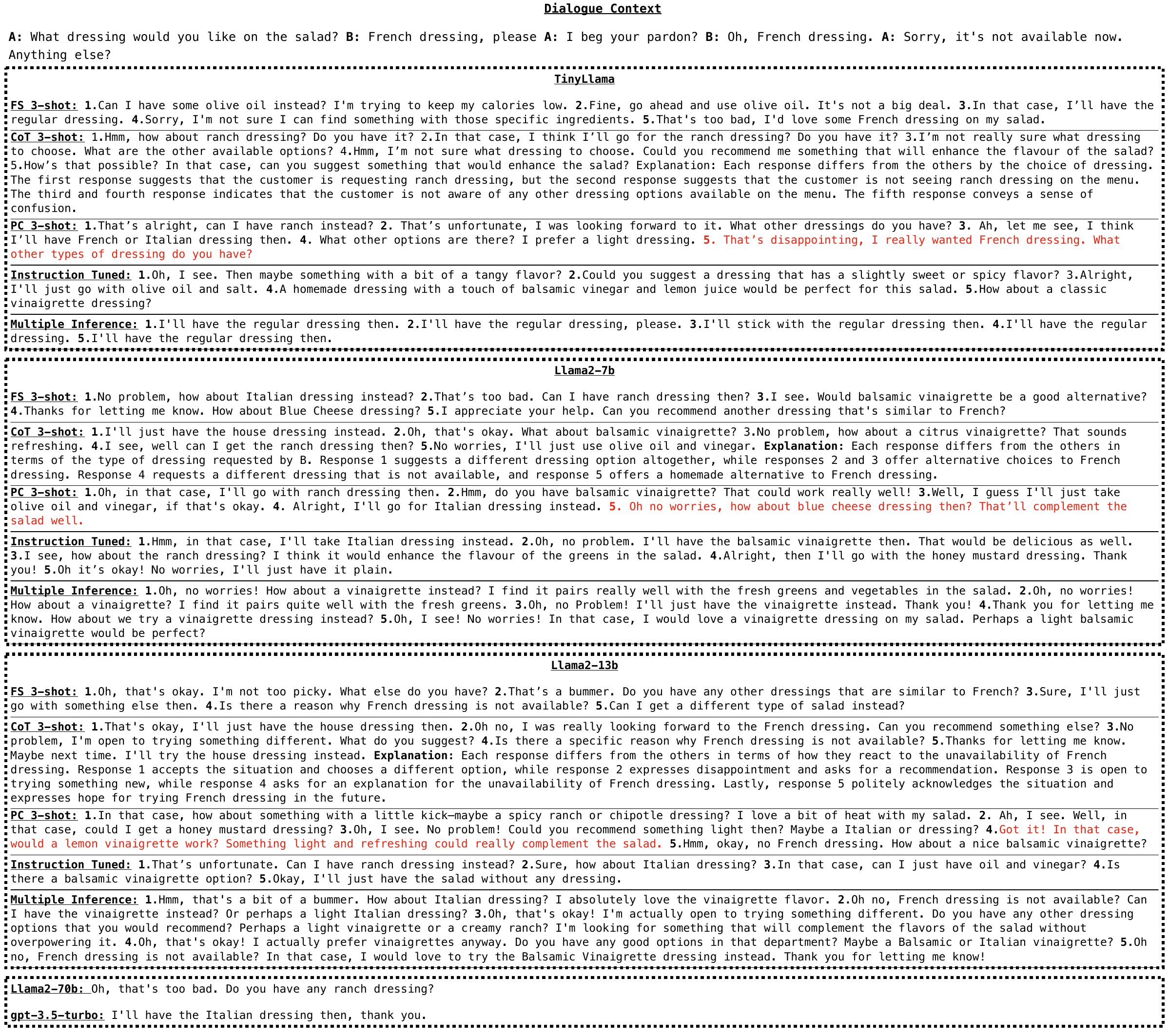}}
    \caption{Samples of response sets generated by TinyLlama, Llama2-7b and Llama2-13b. The responses in \textcolor{red}{red} was selected by the ODRP\textsubscript{HN} model during PS.}
    \label{fig:samples}
\end{figure*}

\end{document}